\documentclass[runningheads,a4paper]{llncs}

\usepackage{times}
\usepackage{amssymb}
\usepackage{graphicx}
\usepackage{url}

\newcommand{\keywords}[1]{\par\addvspace\baselineskip
\noindent\keywordname\enspace\ignorespaces#1}

\usepackage{comment}
\usepackage{multirow}
\usepackage{arydshln}
\usepackage{subcaption}
\usepackage{float}

\begin{document}

% TSD 2025: Put your title here (please, use capitalization, see e.g.
% http://en.wikibooks.org/wiki/Basic_Book_Design/Capitalizing_Words_in_Titles)
\title{An Empirical Analysis of Discrete Unit Representations in Speech Language Modeling Pre-training}

% TSD 2025: a short form should be given in case the title is too long for the running head
\titlerunning{An Empirical Analysis of Discrete Unit Representations in Speech LM Pre-training}

% TSD 2025: Author's names. Chinese authors should write their first names(s)
% in front of their surnames. This ensures that the names appear correctly in
% If the names contain accented characters, please use escape codes
% (refer to http://en.wikibooks.org/wiki/LaTeX/Special_Characters#Escaped_codes)
% TODO: Uncomment for final version
\author{Yanis Labrak\inst{1,3} \and Richard Dufour\inst{2} \and Mickaël Rouvier\inst{1}}

% TSD 2025: For authors from different institutions, please use the following
% form including institution reference
% \author{Firstname1 Surname1\inst{1} \and Firstname2 Surname2 \inst{2}}

% TODO: Uncomment for final version
% TSD 2025: Author's names for headings. For 1-2 authors, use the following form
\authorrunning{Yanis Labrak et al.}
% TSD 2025: For more than 2 authors, please, use the following
% \authorrunning{Name1 Surname1 et al.}

% \affiliation{Laboratoire Informatique d'Avignon}{Avignon University}{Avignon, France}
% \affiliation{Nantes Université}{École Centrale Nantes, CNRS, LS2N, UMR 6004}{Nantes, France}
% \affiliation{}{Zenidoc}{Marseille, France}
% \email{yanis.labrak@univ-avignon.fr, richard.dufour@univ-nantes.fr, mickael.rouvier@univ-avignon.fr}

% TODO: Uncomment for final version
% TSD 2025: The authors' affiliations
\institute{Laboratoire Informatique d'Avignon, Avignon University, Avignon, France \\
Nantes Université, École Centrale Nantes, CNRS, LS2N, UMR 6004, Nantes, France \\
Zenidoc, Marseille, France \\
% TSD 2025: optional url
% \url{www.website.org} \\
% \mailsa\\
% TSD 2025: For authors from different institutions, add 2nd institution, etc.
% \and
% Affiliation2, Institute2, Address \\
% \url{www.website.org} \\
% \mailsb\\
}

% TSD 2025: Put all authors' names to the proceeding index (surname, first name)
% \index{Surname1, Firstname1}
% \index{Surname2, Firstname2}

% TODO: Uncomment for final version 
\index{Labrak, Yanis}
\index{Dufour, Richard}
\index{Rouvier, Mickaël}

\toctitle{} \tocauthor{}

\maketitle

%
%
%	TSD 2025 SUBMISSION TEXT
%
%
\begin{abstract}
% TSD 2025:
This paper investigates discrete unit representations in Speech Language Models (SLMs), focusing on optimizing speech modeling during continual pre-training.
In this paper, we systematically examine how model architecture, data representation, and training robustness influence the pre-training stage in which we adapt existing pre-trained language models to the speech modality.
Our experiments highlight the role of speech encoders and clustering granularity across different model scales, showing how optimal discretization strategies vary with model capacity.
By examining cluster distribution and phonemic alignments, we investigate the effective use of discrete vocabulary, uncovering both linguistic and paralinguistic patterns. Additionally, we explore the impact of clustering data selection on model robustness, highlighting the importance of domain matching between discretization training and target applications.
% TSD 2025: keywords, comma-separated
\end{abstract}

\section{Introduction}

The rapid advancement of pre-trained Large Language Models (LLMs)~\cite{openai2024gpt4technicalreport,touvron2023llama2openfoundation} has transformed Natural Language Processing (NLP), enabling systems with remarkable capabilities in text understanding and generation. However, these models remain largely text-based, overlooking the richness of spoken language, which conveys prosody, emotion, and speaker characteristics essential for human communication.

Recent advances in self-supervised learning have made significant strides toward integrating speech into language modeling. Models such as WavLM~\cite{9814838}, HuBERT~\cite{10.1109/TASLP.2021.3122291} and Wav2Vec 2~\cite{10.5555/3495724.3496768} have proven particularly effective at learning meaningful speech representations without explicit supervision~\cite{10096149}. 

The first attempts to bridge this gap between speech and language modeling emerged with Generative Spoken Language Models (GSLM)~\cite{lakhotia-etal-2021-generative}, demonstrating the possibility of learning directly from raw audio without relying on text supervision. This breakthrough was followed by various approaches to integrate speech into language models~\cite{9864610,nguyen2024spiritlminterleavedspoken}, primarily by incorporating discrete speech representations into their vocabularies~\cite{hassid2024textuallypretrainedspeechlanguage,zhang2023speechgptempoweringlargelanguage,zeng2024glm4voiceintelligenthumanlikeendtoend}.

While speech-extended LLMs have demonstrated promising results in downstream tasks~\cite{cuervo-marxer-2024-scaling,mousavi2024dasbdiscreteaudio}, their performance remains limited~\cite{hassid2024textuallypretrainedspeechlanguage} and the fundamental challenge of optimizing speech modeling during continual pre-training remains largely unexplored. This stage is critical, as it determines how models initially learn to process speech input and serves as the foundation for all subsequent speech-related capabilities~\cite{nguyen2024spiritlminterleavedspoken,maiti2024voxtlmunifieddecoderonlymodels}. 

In this work, we systematically investigate discrete speech unit representations in language modeling. Through extensive experiments across model scales (135M to 1.7B parameters), encoder architectures, and discretization strategies, we address four fundamental questions: the optimal discretization granularity for different encoders, the impact of model scale on semantic information capture, the robustness of speech units to acoustic perturbations, and the nature of linguistic information embedded in these discrete representations.

Our primary contributions are threefold:

\begin{itemize}
    \item First, we trained 51 speech language models of varying capacities (135M to 1.7B parameters) on the spoken language modeling objective introduced by SpeechGPT~\cite{zhang2023speechgptempoweringlargelanguage} through LoRA fine-tuning of pre-trained textual language models of the SmolLM family.
    
    \item Second, we discovered a direct correlation between models' speech modeling capabilities and discrete unit granularity, noting that smaller SLMs struggle to capture semantic information from higher discretization granularity units.
    
    \item Finally, we observed that discrete units' information strongly aligns with phonemes while simultaneously capturing other forms of acoustic information.
\end{itemize}

\begin{figure}[t]
\vskip 0.2in
\begin{center}
\includegraphics[width=0.60\columnwidth]{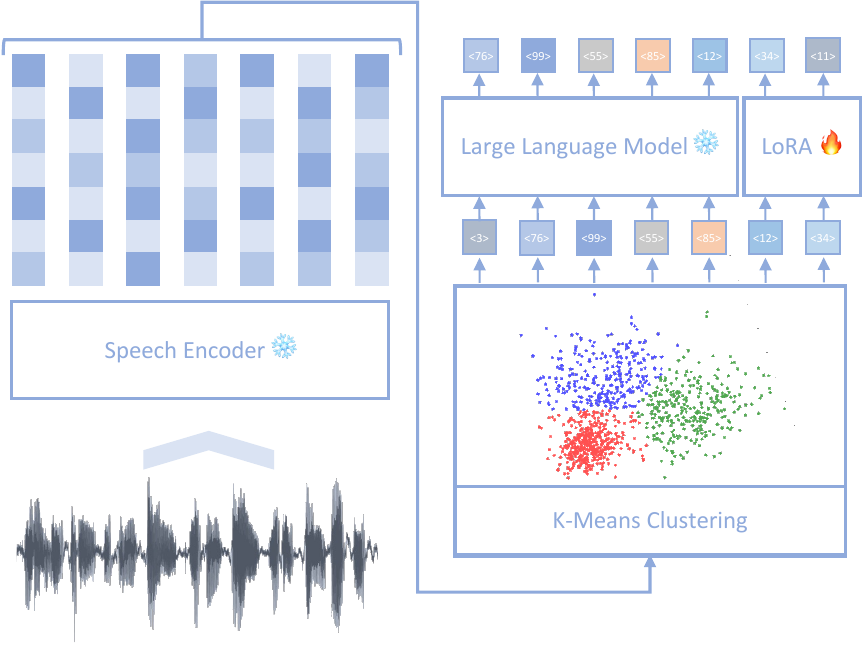}
\caption{Overview of a Speech Language Model.} 
\label{fig:Viz}
\end{center}
\vskip -0.2in
\end{figure}

\section{Spoken Language Modeling}

This section details our methodology for training and comparing our speech-extended language models, with a strong focus on speech representations. The studied SLM architecture follows the approach introduced by SpeechGPT~\cite{zhang2023speechgptempoweringlargelanguage} and relies on discrete units and vocabulary expansion.

% This section describes our approach to extending text language models with speech understanding capabilities, following the methodology introduced by~\cite{zhang2023speechgptempoweringlargelanguage}.

\subsection{Model Architecture}

We experiment with different variants of SmolLM~\cite{benallal2024smollm}, using three model sizes: 135M, 360M, and 1.7B parameters. The core architecture remains unchanged from the original text models, with the only modification being the expansion of the tokenizer vocabulary to incorporate the newer tokens corresponding to the discrete units (see Section~\ref{sec:encodings}).

The training objective follows a standard autoregressive language modeling approach with negative log-likelihood loss. For a sequence of tokens $x = (x_1, ..., x_T)$, the loss is computed as:

\begin{equation}
    \mathcal{L} = -\sum_{t=1}^{T} \log p(x_t|x_{<t})
\end{equation}

where $p(x_t|x_{<t})$ represents the probability of token $x_t$ given all previous tokens in the sequence.
% This loss is applied uniformly whether the tokens represent text or speech units.

Our approach does not aim for full acoustic reconstruction but instead prioritizes semantic modeling of speech as we are focusing on the first stage of speech adaptation of pre-trained textual language models. This stage consists exclusively of learning to process speech units alongside its existing text capabilities, as shown in Figure~\ref{fig:Viz}.

% While lower NLL values indicate better semantic reconstruction ability, they do not necessarily reflect improvements in capturing prosody or paralinguistic features. However, prior work suggests that NLL correlates with semantic understanding tasks~\cite{cuervo-marxer-2024-scaling}, making it a relevant metric for assessing language modeling capabilities in speech-based applications.

Training is conducted on 16 Nvidia H100 80GB GPUs with a batch size of 16 and gradient accumulation of 1. Using a context window of 2,048 tokens, we process 524,288 tokens per step. The training runs for 300 steps, processing approximately 157 million tokens in total. To optimize training efficiency and resource utilization, we incorporate several technical improvements such as LoRA adapters~\cite{hu2021loralowrankadaptationlarge} (rank 64, alpha 16) for parameter-efficient fine-tuning. We chose BFloat16 precision and Flash Attention 2 to reduce memory overhead. It uses AdamW~\cite{loshchilov2018decoupled} optimization with a learning rate of $3\times10^{-4}$ and applies a weight decay coefficient of $0.1$. To ensure reproducible results, the random seed is set to $42$.

\subsection{Speech Encoding and Discretization}
\label{sec:encodings}

To convert the raw speech signal from a continuous form into a discrete one that can be incorporated into the text input of the LLM, we need to have two components: an encoder and a discretizer. Here, we will evaluate four widely used self-supervised speech encoders: WavLM~\cite{9814838}, HuBERT~\cite{10.1109/TASLP.2021.3122291}, XLS-R~\cite{babu22_interspeech}, and  Wav2Vec 2~\cite{10.5555/3495724.3496768}.
For all encoders, we extract features from the final hidden layer, as prior work suggests that this layer provides a strong balance between acoustic and linguistic information~\cite{yang23d_interspeech,9688093}. No additional fine-tuning of the encoders is performed to maintain a fair comparison of their base capabilities.
Each encoder extracts frame-level representations at 50 Hz (20 ms frames), which are then discretized into $k$ clusters that will represent speech units using k-means, following standard practices in spoken language modeling~\cite{zhang2023speechgptempoweringlargelanguage}. To examine the impact of vocabulary size on modeling performance, we experiment with cluster counts of \(k \in \{125, 250, 500, 1000, 2500, 5000\}\).

%$k=\{125,250,500,1000,2500,5000\}$.
% Consecutive identical units are collapsed to reduce redundancy.

The k-means clustering used for speech encoders is trained on 2,000 hours of unlabeled speech for each of the following corpora: LibriHeavy~\cite{kang2023libriheavy}, GigaSpeech~\cite{chen21o_interspeech}, People's Speech~\cite{DBLP:journals/corr/abs-2111-09344}, or CommonVoice 19~\cite{ardila-etal-2020-common}. None of the data selected to build the k-means overlaps with the speech modeling dataset.

% We evaluate four different speech encoders for the discretization of the speech signal: WavLM~\cite{9814838}, HuBERT~\cite{10.1109/TASLP.2021.3122291}, XLS-R~\cite{babu22_interspeech}, and Wav2Vec 2~\cite{10.5555/3495724.3496768}. Each encoder processes raw audio to produce frame-level representations with the same frequency ($20 ms$), which are then quantized into discrete units using k-means clustering. We explore various vocabulary sizes ranging in $k={125,250,500,1000,2500,5000}$ clusters to understand the impact of discretization granularity.

% Following~\cite{labrak84}, we deduplicate consecutive identical speech units to reduce redundancy in the representation while preserving the essential temporal dynamics of the signal. This deduplication step helps maintain a more compact and efficient representation without sacrificing model performance.

\subsection{Speech modeling dataset}

We train the language models to process speech modality using LibriSpeech~\cite{7178964}, a widely used speech corpus containing 960 hours of read English speech. This dataset comprises three subsets (100h, 360h, and 500h), providing a diverse range of speakers and recording conditions. Speech segments are processed through our encoding pipeline (Figure~\ref{fig:Viz}) and using the newly built discrete speech units that serve as input to the language model.

\subsection{Evaluation Methodology}

The effectiveness of each speech unit configuration is measured using Negative Log-Likelihood (NLL) on the LibriSpeech test-clean set. Lower NLL values indicate better modeling of the speech units by the language model, reflecting more stable and predictable representations of the speech signal. Additionally, prior research~\cite{maiti2024voxtlmunifieddecoderonlymodels,cuervo-marxer-2024-scaling} suggests a strong correlation between NLL and performance on semantic speech understanding tasks, such as sWUGGY~\cite{nguyen2020zeroresourcespeechbenchmark}. We maintain consistent frame rates across all models to ensure we can properly compare the NLL. In this case, we use $50\,\mathrm{Hz}$ encoders and a shared tokenizer for all large language models.

% The exact encoders used are:
% - hubert-base-ls960
% - wav2vec2-base-960h
% - wavlm-base-plus
% - wav2vec2-xls-r-300m

%%%%%%%%%%%%%%%%%%%%%%%%%%%%%%%%%%%%%%%%%%
\section{Experiments and results}

We analyze discrete speech units across four dimensions: encoder and discretization methods (Section~\ref{s:compare_enc}), language model scaling (Section~\ref{s:impact_mod}), acoustic robustness (Section~\ref{s:discrete_robust}), and linguistic content (Section~\ref{s:cluster_attrib}).

\subsection{Comparing Encoders and Discretization Granularity}
\label{s:compare_enc}

Results across varying cluster sizes (see Table~\ref{tab:encoder_comparison}) show a consistent initial degradation in performance as the number of clusters increases, with NLL values ranging from 4.2-4.7 ($k=125$) to 7.8-8.1 ($k=5,000$) at Step 100. Training progression significantly improves performance, particularly between Steps 100–200. Among the evaluated encoders, WavLM achieves the best performance (NLL=2.05, $k=500$) at Step 300, followed by smaller cluster configurations ($k=125$, $k=250$), which remain competitive (NLL $\simeq$ 2.15). HuBERT shows similar trends with slightly higher NLL across all cluster sizes, while XLS-R and Wav2Vec consistently underperform, particularly at larger $k$ values.

% The analysis of different speech encoders across varying cluster sizes reveals clear patterns in both convergence and final performance. At the beginning, all encoders exhibit similar behavior, with performance degrading as the number of clusters increases. This degradation is consistent across encoders, with NLL values ranging from approximately 4.2-4.7 at k=125 to 7.8-8.1 at k=5,000.

% Training progression shows substantial improvements across all configurations, with the most significant gains occurring between Steps 100 and 200. WavLM demonstrates superior performance, achieving the best overall score of 2.05 with k=500 at Step 300. This is followed by configurations using k=125 and k=250 clusters, which achieve NLL values of 2.15 and 2.16 respectively. HuBERT shows similar trends but with slightly higher NLL values, while XLS-R and Wav2Vec consistently underperform across all settings.

Notably, smaller cluster sizes ($k \leq 1,000$) consistently yield better performance. In contrast, models using $k \geq 2,500$ experience substantial degradation, with a sharp increase in NLL. This suggests that larger vocabularies introduce excessive speech unit granularity, potentially leading to noisier token distributions and increased token sparsity. As a result, the model struggles to learn stable speech representations, reinforcing the practical advantage of smaller, more compact cluster sets.
% This suggests smaller cluster sizes offer both better performance and practical advantages.

\begin{table}[H]
\centering
\vspace{3mm}
\renewcommand{\arraystretch}{1.0}
\setlength\tabcolsep{5.0pt}
\begin{tabular}{llllll}
\hline
Encoder & Clusters & Step 100 & Step 200 & Step 300 \\
\hline
\multirow{6}{*}{WavLM}
& $k=125$    & 4.681 & 2.502 & 2.149 \\
& $k=250$    & 5.356 & 2.785 & 2.158 \\
& $k=500$    & 6.040 & 2.621 & 2.048 \\
& $k=1,000$  & 6.659 & 3.057 & 2.189 \\
& $k=2,500$  & 7.281 & 5.073 & 4.010 \\
& $k=5,000$  & 7.869 & 5.538 & 4.208 \\
\hdashline
\multirow{6}{*}{HuBERT}
& $k=125$   & 4.705 & 2.596 & 2.240 \\
& $k=250$   & 5.393 & 2.825 & 2.289 \\
& $k=500$   & 6.087 & 2.909 & 2.348 \\
& $k=1,000$ & 6.711 & 3.717 & 2.822 \\
& $k=2,500$ & 7.430 & 4.940 & 3.827 \\
& $k=5,000$ & 8.052 & 5.759 & 4.289 \\
\hdashline
\multirow{6}{*}{XLS-R}
& $k=125$   & 4.205 & 2.694 & 2.433 \\
& $k=250$   & 4.902 & 3.436 & 2.916 \\
& $k=500$   & 5.592 & 3.608 & 3.034 \\
& $k=1,000$ & 6.276 & 3.964 & 3.282 \\
& $k=2,500$ & 7.201 & 5.241 & 4.177 \\
& $k=5,000$ & 7.918 & 6.034 & 4.959 \\
\hdashline
\multirow{6}{*}{Wav2Vec}
& $k=125$   & 4.600 & 3.069 & 2.534 \\
& $k=250$   & 5.153 & 3.559 & 2.880 \\
& $k=500$   & 5.886 & 4.042 & 3.251 \\
& $k=1,000$ & 6.656 & 4.712 & 3.614 \\
& $k=2,500$ & 7.647 & 5.744 & 4.434 \\
& $k=5,000$ & 8.179 & 6.397 & 5.057 \\
\hline
\end{tabular}
\vspace{3mm}
\caption{Negative log likelihood ($\downarrow$) comparison of different encoders with varying cluster sizes and built from 2,000 hours of unlabeled speech from LibriHeavy. Results are reported at training steps 100, 200, and 300.}
\label{tab:encoder_comparison}
\end{table}

% A key finding is the strong correlation between cluster size and model performance. Configurations with fewer clusters ($k\le1,000$) consistently outperform those with larger vocabularies. Performance degrades sharply at $k=2,500$ and $k=5,000$, with NLL values increasing by nearly 2 points across all encoders. This suggests that smaller cluster sizes not only provide better performance but also offer practical advantages through reduced model size and faster inference times.

\subsection{Impact of Model Scale on Discrete Unit Learning}
\label{s:impact_mod}

Table~\ref{tab:encoder_comparison_SmoLM} shows the results obtained with the SmolLM model across different training conditions.
The larger SmolLM-1.7B model significantly outperforms its smaller counterparts, achieving NLL scores of 1.82-1.95 compared to 2.04-2.24 for the 135M model. This improvement suggests that model capacity strongly influences speech unit modeling quality.

% The comparison across different model scales reveals several interesting insights about speech unit modeling. The most striking observation in the Table~\ref{tab:encoder_comparison} is the significant improvement in performance achieved by the larger SmolLM-1.7B model compared to its smaller counterparts, SmolLM-135M and SmolLM-360M. Across most configurations, the 1.7B model consistently achieves better scores, typically around 1.82 and 1.95, compared to 2.14-2.24 for the 135M model. This substantial improvement suggests that model capacity plays a crucial role in the quality of speech unit modeling.

WavLM consistently outperforms HuBERT across all model scales, particularly at lower cluster counts ($k \le 500$). The performance gap between encoders remains relatively stable as model size increases. Larger models show better handling of higher cluster counts, with the 1.7B model demonstrating remarkable stability (NLL 1.83-2.28) within its operational range ($k \le 1,000$), though encountering memory limitations at higher clusters.

Our findings show that the best results are achieved by using larger models with fewer clusters. This approach provides a good balance between model performance and computational efficiency. Additionally, larger models seem to be better at handling both noisy token distributions and sparse token patterns, where smaller models struggle.

\begin{table}[H]
\centering
\vspace{3mm}
\renewcommand{\arraystretch}{1.0}
\setlength\tabcolsep{5.0pt}
\begin{tabular}{ll|ccc}
\hline
& & \multicolumn{3}{c}{SmolLM} \\
Encoder & Clusters & \textit{135M} & \textit{360M} & \textit{1.7B} \\
\hline
\multirow{6}{*}{WavLM}
& $k=125$   & 2.149 & 2.088 & 1.887 \\
& $k=250$   & 2.158 & 2.159 & 1.861 \\
& $k=500$   & 2.048 & 2.210 & 1.829 \\
& $k=1,000$ & 2.189 & 2.386 & 1.937 \\
& $k=2,500$ & 4.010 & 2.674 & \texttt{OOM} \\
& $k=5,000$ & 4.208 & 2.925 & \texttt{OOM} \\
\hdashline
\multirow{6}{*}{HuBERT} 
& $k=125$   & 2.240 & 2.158 & 1.954 \\
& $k=250$   & 2.289 & 2.278 & 2.049 \\
& $k=500$   & 2.348 & 2.499 & 2.137 \\
& $k=1,000$ & 2.822 & 2.698 & 2.282 \\
& $k=2,500$ & 3.827 & 3.054 & \texttt{OOM} \\
& $k=5,000$ & 4.289 & 3.377 & \texttt{OOM} \\
\hline
\end{tabular}
\vspace{3mm}
\caption{Negative log-likelihood ($\downarrow$) comparison of different encoders with varying cluster sizes, built from 2,000 hours of unlabeled speech from LibriHeavy, and trained during 300 steps (approximately 150M tokens).}
\label{tab:encoder_comparison_SmoLM}
\end{table}

\subsection{Discrete Unit Stability Under Audio Perturbations}
\label{s:discrete_robust}

We evaluated discrete unit robustness using a SmolLM-135M model with WavLM encoder ($k=500$) across k-means built from different datasets. Tests included high-intensity Gaussian noise (Noise-H, SNR 15-20dB), low-intensity Gaussian noise (Noise-L, SNR 5-10dB), and random pitch shifts (±5\% range) on the test-clean set of LibriSpeech.

% The robustness of speech units to audio perturbations was evaluated across different training datasets using a SmolLM-135M model with WavLM encoder (k=500). We tested three common perturbations on the test-clean set: high-intensity additive Gaussian noise (Noise-H) with SNR between 15dB and 20dB, low-intensity additive Gaussian noise (Noise-L) with SNR between 5dB and 10dB, and random pitch shifts within a ±5\% range.

\begin{table}[H]
\centering
\renewcommand{\arraystretch}{1.0}
\setlength\tabcolsep{5.0pt}
\vspace{3mm}
\begin{tabular}{l|cccc}
\hline
% & \multicolumn{4}{c}{test-clean} \\
Source k-means & \textit{Clean} & \textit{Noise-H} & \textit{Noise-L} & \textit{Pitch Shift} \\
\hline
LibriHeavy                     & 2.621 & 2.692 & 2.678 & 2.704 \\
GigaSpeech                     & 3.073 & 3.090 & 3.089 & 3.111 \\
People's Speech                   & 2.739 & 2.853 & 2.860 & 2.866 \\
CommonVoice                   & 2.852 & 3.090 & 2.853 & 3.111 \\ 
\hline
\end{tabular}
\vspace{3mm}
\caption{Negative log-likelihood ($\downarrow$) on LibriSpeech test-clean for SmolLM-135M model using WavLM ($k=500$) built from different speech datasets and trained on LibriSpeech during $\approx$1 epoch.}
\label{tab:speech_performance}
\end{table}

% LibriHeavy-trained discrete units show superior performance and stability, with NLL increasing only marginally from 2.62 (clean) to 2.70 (perturbed). Other datasets exhibit higher baseline NLL and greater perturbation sensitivity, with GigaSpeech and CommonVoice showing NLL increases up to 0.26 points. These results suggest domain matching between speech unit training data and target application is crucial for optimal performance and robustness as show by performances on LibriHeavy.

LibriHeavy-trained models show superior performance and stability, with NLL increasing only marginally from 2.62 (clean) to 2.70 (perturbed). Other datasets exhibit higher baseline NLL and greater perturbation sensitivity, with GigaSpeech and CommonVoice showing NLL increases up to 0.26 points. These results suggest that domain matching between speech unit k-means construction data and target application is crucial for optimal performance and robustness, as shown on LibriHeavy. Notably, training on inherently noisy datasets like GigaSpeech and CommonVoice does not improve robustness to perturbations, but rather leads to overall performance degradation, challenging the assumption that exposure to bad acoustic conditions during training necessarily benefits model resilience. Finally, the People's Speech dataset stands out by showing both good overall performance and stability when dealing with noise. This can be attributed to its wide range of audio quality levels and its similarity to the target domain.

% Models trained on LibriHeavy demonstrate the best overall performance and stability, with NLL values ranging from 2.62 on clean audio to 2.70 under perturbations. The relative degradation remains minimal, with only a 0.08 point increase in NLL even under the most challenging conditions. In contrast, models trained on other datasets show both higher baseline NLL and greater sensitivity to perturbations. GigaSpeech and CommonVoice, in particular, exhibit more substantial performance drops under noise and pitch modifications, with NLL increasing by up to 0.26 points (8.31\% relative).

% These results likely reflect the domain similarity between LibriHeavy and the training data (LibriSpeech), rather than inherent dataset quality or diversity. This suggests that matching the domain of the speech unit training data with the target application may be crucial for optimal performance and robustness.

%%%%%%%%%%%%%%%%%%%%%%%%%%%%%%%%%%%%%%%%%
\subsection{Clusters attribution}
\label{s:cluster_attrib}

We analyze cluster usage distribution across encoders and vocabulary sizes to understand their effectiveness in capturing speech phenomena. Using perplexity-based metric:

% Beyond raw performance metrics, understanding how effectively each encoder utilizes its discrete vocabulary provides crucial insights into their behavior and efficiency. To quantify this, we analyze the distribution of cluster usage across different encoders and vocabulary sizes. A well-balanced distribution suggests the encoder is effectively capturing the full range of speech phenomena, while skewed distributions might indicate under-utilization of the available vocabulary space.

\begin{equation}
H_{clusters} = \exp(-\sum_{i=1}^{k} p_i \log p_i)
\end{equation}

where $p_i$ represents each cluster's probability. The resulting value $H_{clusters}$, expressed as a percentage $(\frac{H_{clusters}}{k}) * 100)$, indicates cluster utilization efficiency, with 100\% representing uniform usage.

% This metric computes the effective number of clusters by calculating the perplexity of the cluster distribution. For each cluster $i$, we first compute its probability $p_i$ by dividing its occurrence count $n_i$ by the total number of units. The resulting value $H_{clusters}$ represents how many clusters are effectively being used if they were uniformly distributed. The metric ranges from $1$ (when only one cluster is used) to $k$ (the total number of clusters, when all clusters are used equally). When expressed as a percentage $(\frac{H_{clusters}}{k}) * 100)$, it provides an intuitive measure of cluster utilization, where 100\% indicates perfectly uniform usage and lower percentages suggest more concentrated distributions.

% \vspace{3mm}

% The analysis reveals interesting contrasts between cluster utilization and model performance. HuBERT and WavLM demonstrate superior cluster utilization (77-92\% and 74-91\% respectively) while maintaining strong NLL scores. XLS-R and Wav2Vec show notably lower utilization rates, particularly on the test-other set, with XLS-R dropping to 52-68\% and Wav2Vec ranging from 63-66\% in challenging conditions.

\begin{table}[H]
\centering
\renewcommand{\arraystretch}{1.0}
\setlength\tabcolsep{5.0pt}
%\setlength\extrarowheight{3pt}
% \setlength\extrarowheight{2pt}

% \resizebox{\columnwidth}{!}{
\vspace{3mm}
\begin{tabular}{l|cc|cc|cc|cc}
\hline
& \multicolumn{2}{c|}{$k=250$} & \multicolumn{2}{c|}{$k=1000$} & \multicolumn{2}{c|}{$k=2500$} & \multicolumn{2}{c}{$k=5000$} \\
Model & $C$ & $O$ & $C$ & $O$ & $C$ & $O$ & $C$ & $O$ \\
\hline
WavLM   & 90.9 & 87.3 & 83.8 & 80.3 & 81.8 & 78.5 & 76.5 & 73.9 \\
HuBERT  & 91.9 & 89.9 & 84.5 & 83.2 & 83.3 & 81.1 & 79.7 & 77.6 \\
XLS-R   & 82.5  & 68.0  & 71.4  & 57.7  & 70.3  & 52.1  & 72.4  & 56.0  \\
Wav2Vec & 76.4  & 66.3  & 76.8  & 64.0  & 80.8  & 65.6  & 78.2  & 63.1  \\
\hline
\end{tabular}
\vspace{3mm}
\caption{Cluster utilization percentage (\%) across different models and cluster sizes for test-clean ($C$) and test-other ($O$) sets.}
% }
\label{tab:cluster_utilization}
\end{table}

% \textbf{Practical Implications and Cluster Size Impact}

HuBERT and WavLM demonstrate superior cluster utilization (77-92\% and 74-91\% respectively) while maintaining strong NLL scores, compared to XLS-R (52-68\%) and Wav2Vec (63-66\%). Lower cluster ranges ($k=250$) show optimal utilization across all encoders, with HuBERT and WavLM exceeding 90\% on clean test sets. Comparing test-clean and test-other utilization reveals varying robustness levels. HuBERT and WavLM show minimal degradation (2-4\% drop), while XLS-R and Wav2Vec exhibit larger stability gaps (up to 15-18\% drop) in challenging conditions. This pattern persists across all cluster sizes.
% , though larger vocabularies tend to widen the utilization gap.

% The data suggests an optimal balance at lower cluster ranges ($k=250$), where utilization rates are highest across all encoders. This is particularly evident in the clean test set, where both HuBERT and WavLM achieve over 90\% utilization. For practical applications, HuBERT emerges as the most versatile choice, offering both high cluster utilization and strong robustness, though WavLM's consistency makes it a reliable alternative.

% The results demonstrate varying degrees of robustness across different encoders when comparing test-clean and test-other performance. HuBERT exhibits the strongest robustness with minimal degradation between clean and other conditions (2-3\% drop), followed closely by WavLM (3-4\% drop). In contrast, XLS-R and Wav2Vec show significantly larger stability gaps, with utilization dropping by up to 18\% and 15\% respectively in challenging acoustic conditions. This pattern is consistent across all cluster sizes, though the gap tends to widen with larger vocabularies.

%%%%%%%%%%%%%%%%%%%%%%%%%%%%%%%%%%%%%%%%%
\subsection{Discrete unit alignment with phonemes}

\begin{figure}[H]
    \centering
    \begin{minipage}{\textwidth}
        \begin{subfigure}[b]{0.46\textwidth}
            \includegraphics[width=\textwidth]{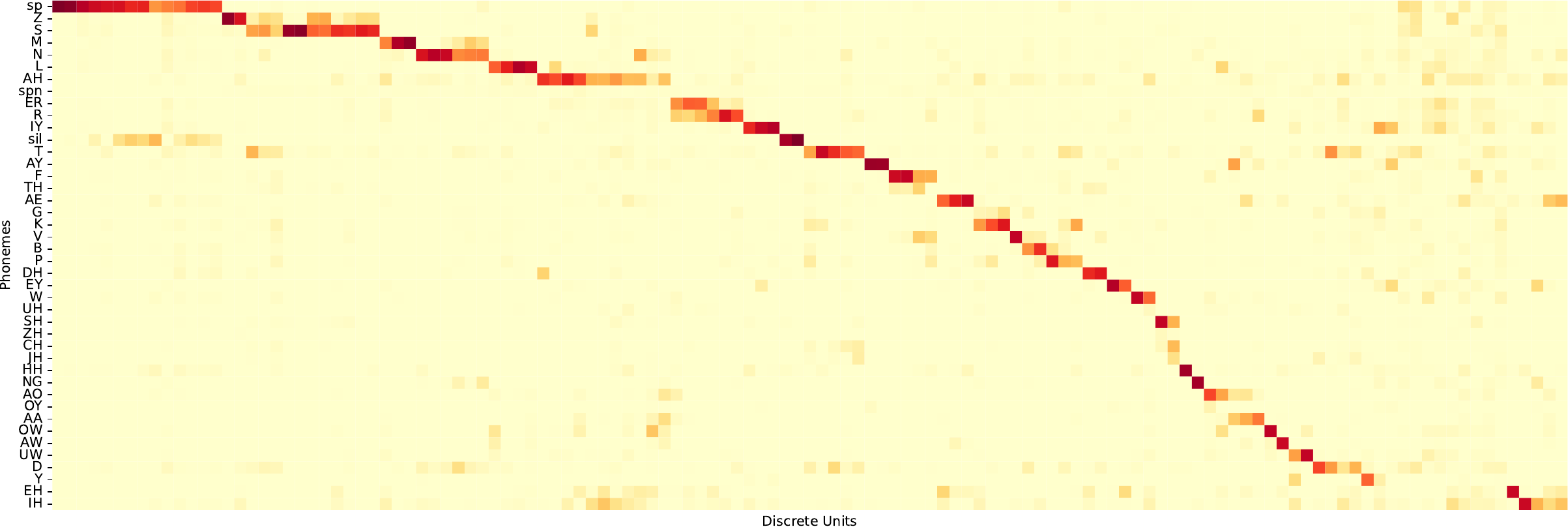}
            \caption{Discrete units trained on GigaSpeech}
        \end{subfigure}
        \hspace{0.01\textwidth}
        \begin{subfigure}[b]{0.46\textwidth}
            \includegraphics[width=\textwidth]{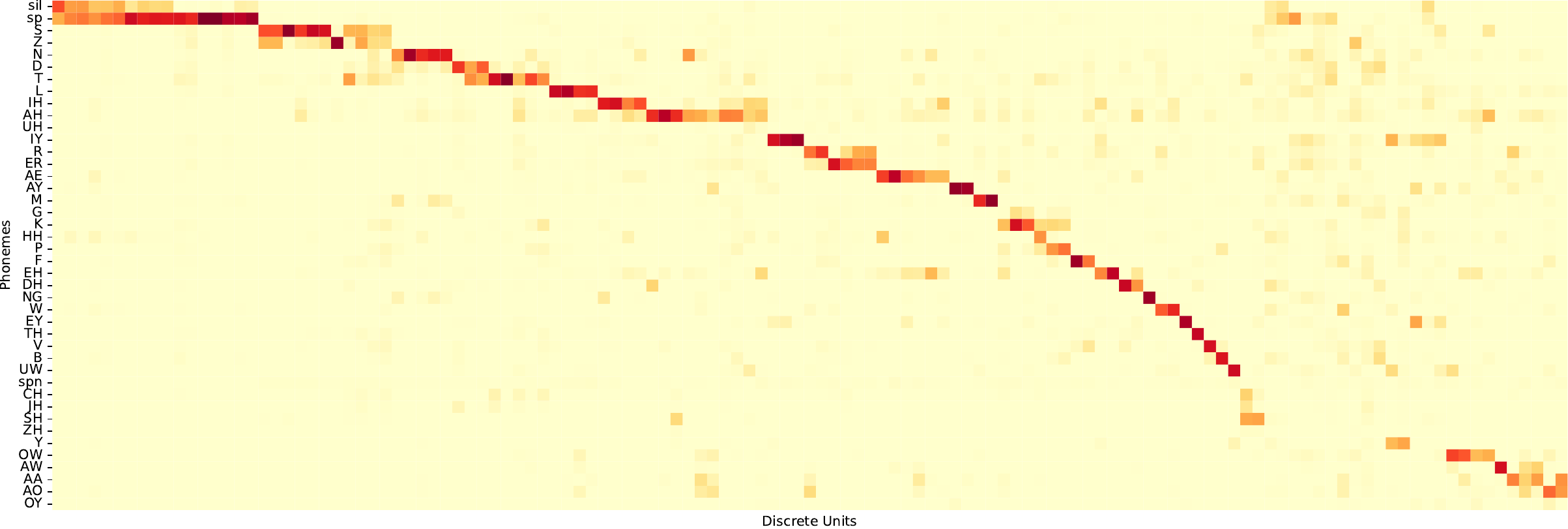}
            \caption{Discrete units trained on People's Speech}
        \end{subfigure}
        
        \vspace{0.5cm}
        \begin{subfigure}[b]{0.46\textwidth}
            \includegraphics[width=\textwidth]{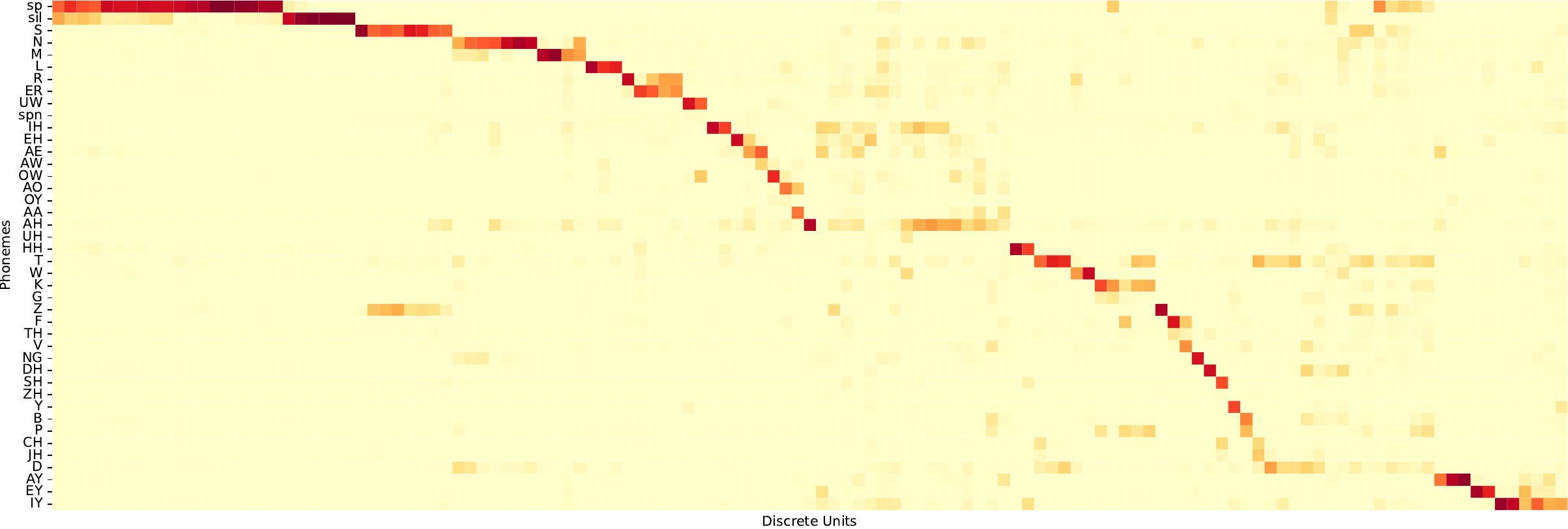}
            \caption{Discrete units trained on CommonVoice}
        \end{subfigure}
        \hspace{0.01\textwidth}
        \begin{subfigure}[b]{0.46\textwidth}
            \includegraphics[width=\textwidth]{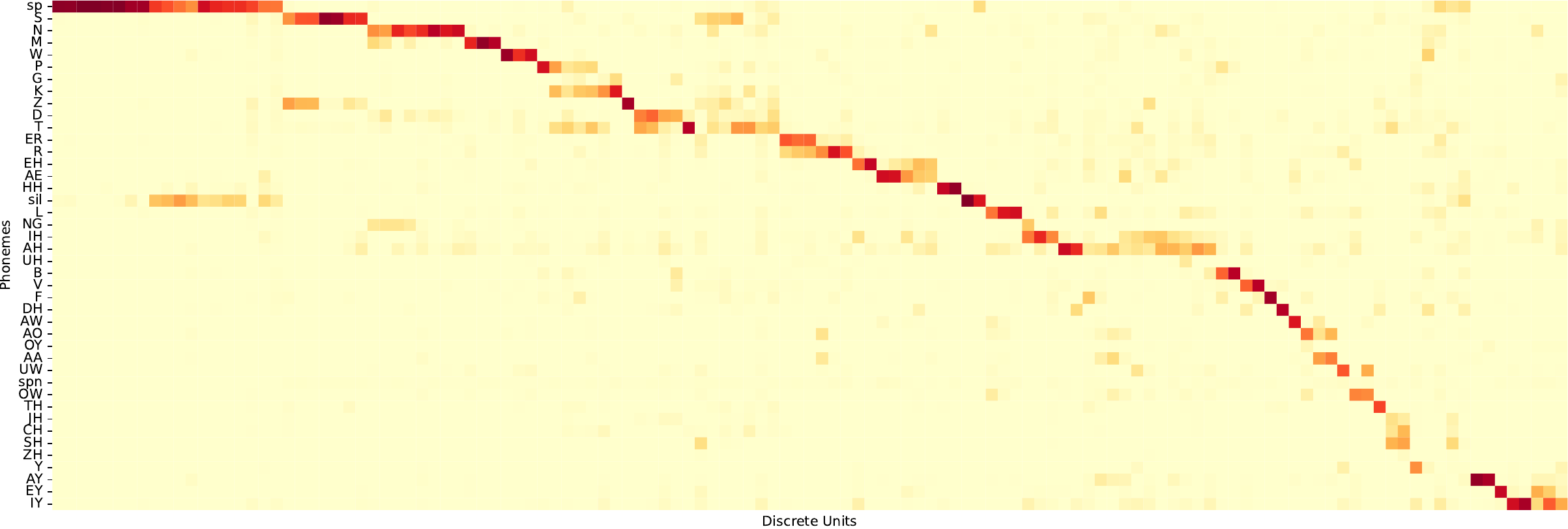}
            \caption{Discrete units trained on LibriHeavy}
        \end{subfigure}
    \end{minipage}
    \vspace{3mm}
    \caption{Phoneme confusion matrices showing the relationship between predicted discrete units and ground truth phonemes. Each matrix represents discrete units k-means built from a different dataset. All of them are based on WavLM ($k=125$) and represent LibriSpeech test-clean subset.}
    \label{fig:confusion_matrix_phonemes_units}
\end{figure}

To better understand what discrete units represent and to try to understand if they capture phonetic information, we analyze their alignment with phonemes using forced alignment from the Montreal Forced Aligner (MFA)~\cite{mcauliffe17_interspeech} on LibriSpeech test clean. We compute for each discrete unit its temporal overlap with the aligned phonemes, creating a probability distribution over phonemes for each unit. Figure~\ref{fig:confusion_matrix_phonemes_units} visualizes this alignment as a matrix where rows represent phonemes and columns represent discrete units, with color intensity indicating the probability of association. The clear diagonal pattern reveals that discrete units learn to specialize in specific phonemes, suggesting the model has captured meaningful phonetic structure. This specialization is particularly strong for distinctive phonemes like vowels (\texttt{/AH/}, \texttt{/IY/}, \texttt{/UW/}), certain consonants (\texttt{/S/}, \texttt{/F/}, \texttt{/M/}) and silence, which show dark regions of high probability along the diagonal for a few sets of units.

Interestingly, we observe some natural clustering of acoustically similar phonemes. For instance, related vowel sounds tend to share similar units, as do phonetically similar consonants. This suggests the discretization process captures not just individual phonemes but also underlying phonetic features. The sparse off-diagonal elements indicate minimal confusion between dissimilar phonemes, demonstrating the model's ability to learn discriminative representations.

\begin{figure}[H]
    \centering
    \begin{minipage}{\textwidth}
        \begin{subfigure}[b]{1.0\textwidth}
            \includegraphics[width=\textwidth]{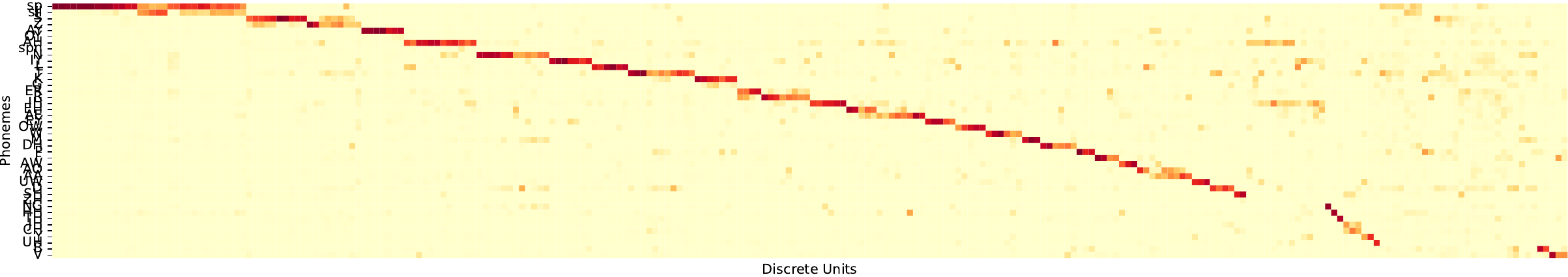}
            \caption{Discrete units trained on People's Speech Test Clean with 250 WavLM clusters.}
        \end{subfigure}
        
        \vspace{0.5cm}
        
        \begin{subfigure}[b]{1.0\textwidth}
            \includegraphics[width=\textwidth]{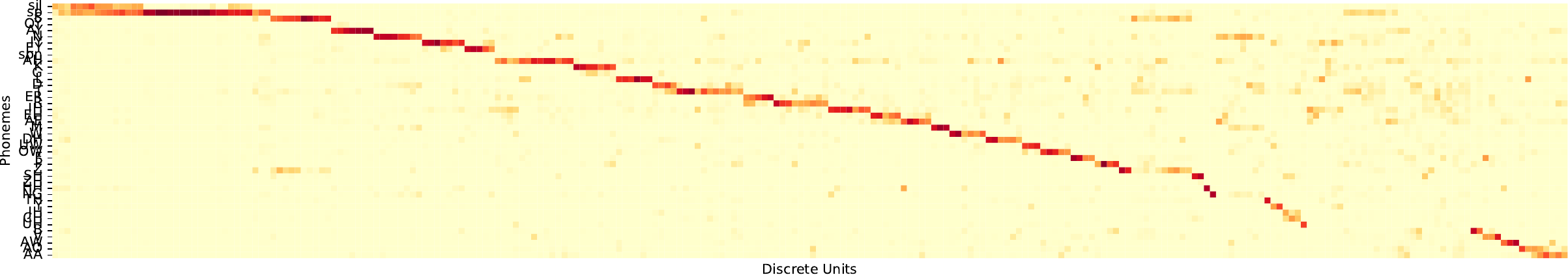}
            \caption{Discrete units trained on People's Speech Test Other with 250 WavLM clusters.}
        \end{subfigure}
    \end{minipage}
    \caption{Phoneme confusion matrices showing the relationship between predicted discrete units and ground truth phonemes. Each matrix represents discrete units k-means built from a different dataset. All of them are based on WavLM ($k=250$) and represent LibriSpeech test-clean subset.}
    \label{fig:confusion_matrix_phonemes_units_larger}
\end{figure}

The alignment quality remains consistent across different k-means building sources and shows a similar pattern across all the granularities (see Figure~\ref{fig:confusion_matrix_phonemes_units_larger}), but wasn't displayed due to a lack of space. This analysis provides quantitative evidence that self-supervised discrete units can effectively capture phoneme-level distinctions without explicit phonetic supervision, supporting their use as intermediate representations for speech processing tasks.

When we increase the number of clusters such as in the Figure~\ref{fig:confusion_matrix_phonemes_units_larger}, similar phonetic patterns remain clearly visible, with the diagonal structure preserved but becoming more fine-grained. The higher cluster count (250) allows for more specialized unit-to-phoneme mappings while maintaining the overall phonetic organization. This suggests that even at higher granularity, discrete units continue to capture meaningful phonetic distinctions, with each phoneme being represented by a more specific set of units rather than becoming fragmented across unrelated regions.

\section{Conclusion}

This work presents a comprehensive empirical analysis of discrete unit representations in speech language modeling, providing key insights into their behavior and optimization at the pre-training stage. Through extensive experiments across model scales and encoder architectures, we demonstrate that smaller discrete vocabularies ($k \le 1,000$) consistently achieve superior performance, with WavLM-based units showing particular promise. The relationship between model scale and unit learning reveals that larger models (1.7B parameters) exhibit enhanced robustness to vocabulary size and better handle acoustic variations, suggesting more abstract speech representation learning.

Our analysis of cluster utilization and phonemic alignments demonstrates that self-supervised discrete units naturally capture phonetic structure without explicit supervision. The strong correlation between domain matching and model performance, particularly evident in LibriHeavy-trained units, emphasizes the importance of careful data selection for discrete unit training.

These findings have important implications for the design of speech adaptation of existing pre-trained large language models, suggesting that optimal performance may be achieved through a combination of moderate vocabulary sizes, domain-matched training data, and sufficient model capacity. We release our tokenized datasets and clustering models on GitHub and HuggingFace to facilitate further research in this direction.

% Several promising research directions emerge from this work. First, for resource-constrained environments like on-device E2E systems, investigating intelligent vocabulary reduction through phoneme-based unit merging could improve efficiency and reduce memory footprint associated with speech integration. However, this requires careful consideration of the trade-off between compression and preservation of context-dependent phonetic variations. Second, inspired by~\cite{zhang2024speechtokenizer}, exploring hierarchical discrete representations could enable simultaneous capture of both fine-grained acoustic details and broader phonetic patterns.
Finally, understanding the balance between semantic and paralinguistic information remains crucial, necessitating evaluation across diverse tasks including Spoken Question Answering, Spoken Language Understanding, and ASR.

\section{Acknowledgements}

This work was performed using HPC resources from GENCI-IDRIS (Grant AD011013061R3 and A0161014871). This work was financially supported by ANR MALADES (ANR-23-IAS1-0005), BPI PARTAGES and Zenidoc.

% Bibliography
\bibliographystyle{splncs04}
\bibliography{paper}

\end{document}